\documentclass[letterpaper,10pt,conference]{cssconf}
\pdfoutput=1
\usepackage[draft]{minted}
\usepackage{graphics} 
\usepackage{epsfig} 
\usepackage{mathptmx} 
\usepackage{times} 
\usepackage{amsmath} 
\usepackage{amssymb}  
\usepackage{subfig}
\usepackage{pgf}
\usepackage{pgfplots,pgfplotstable} 
\usepgfplotslibrary{units}
\title{\LARGE \bf
A machine learning approach to reconstruction of heart surface
potentials from body surface potentials
}

\author{Avinash Malik$^{1}$, Tommy Peng$^{1}$ and Mark Trew$^{2}$
  \thanks{$^{1}$Avinash Malik and Tommy Peng are with Faculty of
    Engineering, University of Auckland, New Zealand. {\tt\small
      avinash.malik@auckland.ac.nz}}%
  \thanks{$^{2}$Mark Trew is with the Auckland Bioengineering Institute,
    Auckland, New Zealand.
    {\tt\small b.d.researcher@ieee.org}}%
}

\makeatletter
\def\PYG@reset{\let\PYG@it=\relax \let\PYG@bf=\relax%
    \let\PYG@ul=\relax \let\PYG@tc=\relax%
    \let\PYG@bc=\relax \let\PYG@ff=\relax}
\def\PYG@tok#1{\csname PYG@tok@#1\endcsname}
\def\PYG@toks#1+{\ifx\relax#1\empty\else%
    \PYG@tok{#1}\expandafter\PYG@toks\fi}
\def\PYG@do#1{\PYG@bc{\PYG@tc{\PYG@ul{%
    \PYG@it{\PYG@bf{\PYG@ff{#1}}}}}}}
\def\PYG#1#2{\PYG@reset\PYG@toks#1+\relax+\PYG@do{#2}}

\expandafter\def\csname PYG@tok@w\endcsname{\def\PYG@tc##1{\textcolor[rgb]{0.73,0.73,0.73}{##1}}}
\expandafter\def\csname PYG@tok@c\endcsname{\let\PYG@it=\textit\def\PYG@tc##1{\textcolor[rgb]{0.25,0.50,0.50}{##1}}}
\expandafter\def\csname PYG@tok@cp\endcsname{\def\PYG@tc##1{\textcolor[rgb]{0.74,0.48,0.00}{##1}}}
\expandafter\def\csname PYG@tok@k\endcsname{\let\PYG@bf=\textbf\def\PYG@tc##1{\textcolor[rgb]{0.00,0.50,0.00}{##1}}}
\expandafter\def\csname PYG@tok@kp\endcsname{\def\PYG@tc##1{\textcolor[rgb]{0.00,0.50,0.00}{##1}}}
\expandafter\def\csname PYG@tok@kt\endcsname{\def\PYG@tc##1{\textcolor[rgb]{0.69,0.00,0.25}{##1}}}
\expandafter\def\csname PYG@tok@o\endcsname{\def\PYG@tc##1{\textcolor[rgb]{0.40,0.40,0.40}{##1}}}
\expandafter\def\csname PYG@tok@ow\endcsname{\let\PYG@bf=\textbf\def\PYG@tc##1{\textcolor[rgb]{0.67,0.13,1.00}{##1}}}
\expandafter\def\csname PYG@tok@nb\endcsname{\def\PYG@tc##1{\textcolor[rgb]{0.00,0.50,0.00}{##1}}}
\expandafter\def\csname PYG@tok@nf\endcsname{\def\PYG@tc##1{\textcolor[rgb]{0.00,0.00,1.00}{##1}}}
\expandafter\def\csname PYG@tok@nc\endcsname{\let\PYG@bf=\textbf\def\PYG@tc##1{\textcolor[rgb]{0.00,0.00,1.00}{##1}}}
\expandafter\def\csname PYG@tok@nn\endcsname{\let\PYG@bf=\textbf\def\PYG@tc##1{\textcolor[rgb]{0.00,0.00,1.00}{##1}}}
\expandafter\def\csname PYG@tok@ne\endcsname{\let\PYG@bf=\textbf\def\PYG@tc##1{\textcolor[rgb]{0.82,0.25,0.23}{##1}}}
\expandafter\def\csname PYG@tok@nv\endcsname{\def\PYG@tc##1{\textcolor[rgb]{0.10,0.09,0.49}{##1}}}
\expandafter\def\csname PYG@tok@no\endcsname{\def\PYG@tc##1{\textcolor[rgb]{0.53,0.00,0.00}{##1}}}
\expandafter\def\csname PYG@tok@nl\endcsname{\def\PYG@tc##1{\textcolor[rgb]{0.63,0.63,0.00}{##1}}}
\expandafter\def\csname PYG@tok@ni\endcsname{\let\PYG@bf=\textbf\def\PYG@tc##1{\textcolor[rgb]{0.60,0.60,0.60}{##1}}}
\expandafter\def\csname PYG@tok@na\endcsname{\def\PYG@tc##1{\textcolor[rgb]{0.49,0.56,0.16}{##1}}}
\expandafter\def\csname PYG@tok@nt\endcsname{\let\PYG@bf=\textbf\def\PYG@tc##1{\textcolor[rgb]{0.00,0.50,0.00}{##1}}}
\expandafter\def\csname PYG@tok@nd\endcsname{\def\PYG@tc##1{\textcolor[rgb]{0.67,0.13,1.00}{##1}}}
\expandafter\def\csname PYG@tok@s\endcsname{\def\PYG@tc##1{\textcolor[rgb]{0.73,0.13,0.13}{##1}}}
\expandafter\def\csname PYG@tok@sd\endcsname{\let\PYG@it=\textit\def\PYG@tc##1{\textcolor[rgb]{0.73,0.13,0.13}{##1}}}
\expandafter\def\csname PYG@tok@si\endcsname{\let\PYG@bf=\textbf\def\PYG@tc##1{\textcolor[rgb]{0.73,0.40,0.53}{##1}}}
\expandafter\def\csname PYG@tok@se\endcsname{\let\PYG@bf=\textbf\def\PYG@tc##1{\textcolor[rgb]{0.73,0.40,0.13}{##1}}}
\expandafter\def\csname PYG@tok@sr\endcsname{\def\PYG@tc##1{\textcolor[rgb]{0.73,0.40,0.53}{##1}}}
\expandafter\def\csname PYG@tok@ss\endcsname{\def\PYG@tc##1{\textcolor[rgb]{0.10,0.09,0.49}{##1}}}
\expandafter\def\csname PYG@tok@sx\endcsname{\def\PYG@tc##1{\textcolor[rgb]{0.00,0.50,0.00}{##1}}}
\expandafter\def\csname PYG@tok@m\endcsname{\def\PYG@tc##1{\textcolor[rgb]{0.40,0.40,0.40}{##1}}}
\expandafter\def\csname PYG@tok@gh\endcsname{\let\PYG@bf=\textbf\def\PYG@tc##1{\textcolor[rgb]{0.00,0.00,0.50}{##1}}}
\expandafter\def\csname PYG@tok@gu\endcsname{\let\PYG@bf=\textbf\def\PYG@tc##1{\textcolor[rgb]{0.50,0.00,0.50}{##1}}}
\expandafter\def\csname PYG@tok@gd\endcsname{\def\PYG@tc##1{\textcolor[rgb]{0.63,0.00,0.00}{##1}}}
\expandafter\def\csname PYG@tok@gi\endcsname{\def\PYG@tc##1{\textcolor[rgb]{0.00,0.63,0.00}{##1}}}
\expandafter\def\csname PYG@tok@gr\endcsname{\def\PYG@tc##1{\textcolor[rgb]{1.00,0.00,0.00}{##1}}}
\expandafter\def\csname PYG@tok@ge\endcsname{\let\PYG@it=\textit}
\expandafter\def\csname PYG@tok@gs\endcsname{\let\PYG@bf=\textbf}
\expandafter\def\csname PYG@tok@gp\endcsname{\let\PYG@bf=\textbf\def\PYG@tc##1{\textcolor[rgb]{0.00,0.00,0.50}{##1}}}
\expandafter\def\csname PYG@tok@go\endcsname{\def\PYG@tc##1{\textcolor[rgb]{0.53,0.53,0.53}{##1}}}
\expandafter\def\csname PYG@tok@gt\endcsname{\def\PYG@tc##1{\textcolor[rgb]{0.00,0.27,0.87}{##1}}}
\expandafter\def\csname PYG@tok@err\endcsname{\def\PYG@bc##1{\setlength{\fboxsep}{0pt}\fcolorbox[rgb]{1.00,0.00,0.00}{1,1,1}{\strut ##1}}}
\expandafter\def\csname PYG@tok@kc\endcsname{\let\PYG@bf=\textbf\def\PYG@tc##1{\textcolor[rgb]{0.00,0.50,0.00}{##1}}}
\expandafter\def\csname PYG@tok@kd\endcsname{\let\PYG@bf=\textbf\def\PYG@tc##1{\textcolor[rgb]{0.00,0.50,0.00}{##1}}}
\expandafter\def\csname PYG@tok@kn\endcsname{\let\PYG@bf=\textbf\def\PYG@tc##1{\textcolor[rgb]{0.00,0.50,0.00}{##1}}}
\expandafter\def\csname PYG@tok@kr\endcsname{\let\PYG@bf=\textbf\def\PYG@tc##1{\textcolor[rgb]{0.00,0.50,0.00}{##1}}}
\expandafter\def\csname PYG@tok@bp\endcsname{\def\PYG@tc##1{\textcolor[rgb]{0.00,0.50,0.00}{##1}}}
\expandafter\def\csname PYG@tok@fm\endcsname{\def\PYG@tc##1{\textcolor[rgb]{0.00,0.00,1.00}{##1}}}
\expandafter\def\csname PYG@tok@vc\endcsname{\def\PYG@tc##1{\textcolor[rgb]{0.10,0.09,0.49}{##1}}}
\expandafter\def\csname PYG@tok@vg\endcsname{\def\PYG@tc##1{\textcolor[rgb]{0.10,0.09,0.49}{##1}}}
\expandafter\def\csname PYG@tok@vi\endcsname{\def\PYG@tc##1{\textcolor[rgb]{0.10,0.09,0.49}{##1}}}
\expandafter\def\csname PYG@tok@vm\endcsname{\def\PYG@tc##1{\textcolor[rgb]{0.10,0.09,0.49}{##1}}}
\expandafter\def\csname PYG@tok@sa\endcsname{\def\PYG@tc##1{\textcolor[rgb]{0.73,0.13,0.13}{##1}}}
\expandafter\def\csname PYG@tok@sb\endcsname{\def\PYG@tc##1{\textcolor[rgb]{0.73,0.13,0.13}{##1}}}
\expandafter\def\csname PYG@tok@sc\endcsname{\def\PYG@tc##1{\textcolor[rgb]{0.73,0.13,0.13}{##1}}}
\expandafter\def\csname PYG@tok@dl\endcsname{\def\PYG@tc##1{\textcolor[rgb]{0.73,0.13,0.13}{##1}}}
\expandafter\def\csname PYG@tok@s2\endcsname{\def\PYG@tc##1{\textcolor[rgb]{0.73,0.13,0.13}{##1}}}
\expandafter\def\csname PYG@tok@sh\endcsname{\def\PYG@tc##1{\textcolor[rgb]{0.73,0.13,0.13}{##1}}}
\expandafter\def\csname PYG@tok@s1\endcsname{\def\PYG@tc##1{\textcolor[rgb]{0.73,0.13,0.13}{##1}}}
\expandafter\def\csname PYG@tok@mb\endcsname{\def\PYG@tc##1{\textcolor[rgb]{0.40,0.40,0.40}{##1}}}
\expandafter\def\csname PYG@tok@mf\endcsname{\def\PYG@tc##1{\textcolor[rgb]{0.40,0.40,0.40}{##1}}}
\expandafter\def\csname PYG@tok@mh\endcsname{\def\PYG@tc##1{\textcolor[rgb]{0.40,0.40,0.40}{##1}}}
\expandafter\def\csname PYG@tok@mi\endcsname{\def\PYG@tc##1{\textcolor[rgb]{0.40,0.40,0.40}{##1}}}
\expandafter\def\csname PYG@tok@il\endcsname{\def\PYG@tc##1{\textcolor[rgb]{0.40,0.40,0.40}{##1}}}
\expandafter\def\csname PYG@tok@mo\endcsname{\def\PYG@tc##1{\textcolor[rgb]{0.40,0.40,0.40}{##1}}}
\expandafter\def\csname PYG@tok@ch\endcsname{\let\PYG@it=\textit\def\PYG@tc##1{\textcolor[rgb]{0.25,0.50,0.50}{##1}}}
\expandafter\def\csname PYG@tok@cm\endcsname{\let\PYG@it=\textit\def\PYG@tc##1{\textcolor[rgb]{0.25,0.50,0.50}{##1}}}
\expandafter\def\csname PYG@tok@cpf\endcsname{\let\PYG@it=\textit\def\PYG@tc##1{\textcolor[rgb]{0.25,0.50,0.50}{##1}}}
\expandafter\def\csname PYG@tok@c1\endcsname{\let\PYG@it=\textit\def\PYG@tc##1{\textcolor[rgb]{0.25,0.50,0.50}{##1}}}
\expandafter\def\csname PYG@tok@cs\endcsname{\let\PYG@it=\textit\def\PYG@tc##1{\textcolor[rgb]{0.25,0.50,0.50}{##1}}}

\def\PYGZus{\char`\_}

\def\PYGZgt{\char`\>}

\def\PYGZpc{\char`\%}

\def\PYGZhy{\char`\-}
\def\PYGZsq{\char`\'}

\def\PYGZti{\char`\~}

\makeatother

\makeatletter
\def\PYGdefault@reset{\let\PYGdefault@it=\relax \let\PYGdefault@bf=\relax%
    \let\PYGdefault@ul=\relax \let\PYGdefault@tc=\relax%
    \let\PYGdefault@bc=\relax \let\PYGdefault@ff=\relax}
\def\PYGdefault@tok#1{\csname PYGdefault@tok@#1\endcsname}
\def\PYGdefault@toks#1+{\ifx\relax#1\empty\else%
    \PYGdefault@tok{#1}\expandafter\PYGdefault@toks\fi}
\def\PYGdefault@do#1{\PYGdefault@bc{\PYGdefault@tc{\PYGdefault@ul{%
    \PYGdefault@it{\PYGdefault@bf{\PYGdefault@ff{#1}}}}}}}
\def\PYGdefault#1#2{\PYGdefault@reset\PYGdefault@toks#1+\relax+\PYGdefault@do{#2}}

\expandafter\def\csname PYGdefault@tok@w\endcsname{\def\PYGdefault@tc##1{\textcolor[rgb]{0.73,0.73,0.73}{##1}}}
\expandafter\def\csname PYGdefault@tok@c\endcsname{\let\PYGdefault@it=\textit\def\PYGdefault@tc##1{\textcolor[rgb]{0.25,0.50,0.50}{##1}}}
\expandafter\def\csname PYGdefault@tok@cp\endcsname{\def\PYGdefault@tc##1{\textcolor[rgb]{0.74,0.48,0.00}{##1}}}
\expandafter\def\csname PYGdefault@tok@k\endcsname{\let\PYGdefault@bf=\textbf\def\PYGdefault@tc##1{\textcolor[rgb]{0.00,0.50,0.00}{##1}}}
\expandafter\def\csname PYGdefault@tok@kp\endcsname{\def\PYGdefault@tc##1{\textcolor[rgb]{0.00,0.50,0.00}{##1}}}
\expandafter\def\csname PYGdefault@tok@kt\endcsname{\def\PYGdefault@tc##1{\textcolor[rgb]{0.69,0.00,0.25}{##1}}}
\expandafter\def\csname PYGdefault@tok@o\endcsname{\def\PYGdefault@tc##1{\textcolor[rgb]{0.40,0.40,0.40}{##1}}}
\expandafter\def\csname PYGdefault@tok@ow\endcsname{\let\PYGdefault@bf=\textbf\def\PYGdefault@tc##1{\textcolor[rgb]{0.67,0.13,1.00}{##1}}}
\expandafter\def\csname PYGdefault@tok@nb\endcsname{\def\PYGdefault@tc##1{\textcolor[rgb]{0.00,0.50,0.00}{##1}}}
\expandafter\def\csname PYGdefault@tok@nf\endcsname{\def\PYGdefault@tc##1{\textcolor[rgb]{0.00,0.00,1.00}{##1}}}
\expandafter\def\csname PYGdefault@tok@nc\endcsname{\let\PYGdefault@bf=\textbf\def\PYGdefault@tc##1{\textcolor[rgb]{0.00,0.00,1.00}{##1}}}
\expandafter\def\csname PYGdefault@tok@nn\endcsname{\let\PYGdefault@bf=\textbf\def\PYGdefault@tc##1{\textcolor[rgb]{0.00,0.00,1.00}{##1}}}
\expandafter\def\csname PYGdefault@tok@ne\endcsname{\let\PYGdefault@bf=\textbf\def\PYGdefault@tc##1{\textcolor[rgb]{0.82,0.25,0.23}{##1}}}
\expandafter\def\csname PYGdefault@tok@nv\endcsname{\def\PYGdefault@tc##1{\textcolor[rgb]{0.10,0.09,0.49}{##1}}}
\expandafter\def\csname PYGdefault@tok@no\endcsname{\def\PYGdefault@tc##1{\textcolor[rgb]{0.53,0.00,0.00}{##1}}}
\expandafter\def\csname PYGdefault@tok@nl\endcsname{\def\PYGdefault@tc##1{\textcolor[rgb]{0.63,0.63,0.00}{##1}}}
\expandafter\def\csname PYGdefault@tok@ni\endcsname{\let\PYGdefault@bf=\textbf\def\PYGdefault@tc##1{\textcolor[rgb]{0.60,0.60,0.60}{##1}}}
\expandafter\def\csname PYGdefault@tok@na\endcsname{\def\PYGdefault@tc##1{\textcolor[rgb]{0.49,0.56,0.16}{##1}}}
\expandafter\def\csname PYGdefault@tok@nt\endcsname{\let\PYGdefault@bf=\textbf\def\PYGdefault@tc##1{\textcolor[rgb]{0.00,0.50,0.00}{##1}}}
\expandafter\def\csname PYGdefault@tok@nd\endcsname{\def\PYGdefault@tc##1{\textcolor[rgb]{0.67,0.13,1.00}{##1}}}
\expandafter\def\csname PYGdefault@tok@s\endcsname{\def\PYGdefault@tc##1{\textcolor[rgb]{0.73,0.13,0.13}{##1}}}
\expandafter\def\csname PYGdefault@tok@sd\endcsname{\let\PYGdefault@it=\textit\def\PYGdefault@tc##1{\textcolor[rgb]{0.73,0.13,0.13}{##1}}}
\expandafter\def\csname PYGdefault@tok@si\endcsname{\let\PYGdefault@bf=\textbf\def\PYGdefault@tc##1{\textcolor[rgb]{0.73,0.40,0.53}{##1}}}
\expandafter\def\csname PYGdefault@tok@se\endcsname{\let\PYGdefault@bf=\textbf\def\PYGdefault@tc##1{\textcolor[rgb]{0.73,0.40,0.13}{##1}}}
\expandafter\def\csname PYGdefault@tok@sr\endcsname{\def\PYGdefault@tc##1{\textcolor[rgb]{0.73,0.40,0.53}{##1}}}
\expandafter\def\csname PYGdefault@tok@ss\endcsname{\def\PYGdefault@tc##1{\textcolor[rgb]{0.10,0.09,0.49}{##1}}}
\expandafter\def\csname PYGdefault@tok@sx\endcsname{\def\PYGdefault@tc##1{\textcolor[rgb]{0.00,0.50,0.00}{##1}}}
\expandafter\def\csname PYGdefault@tok@m\endcsname{\def\PYGdefault@tc##1{\textcolor[rgb]{0.40,0.40,0.40}{##1}}}
\expandafter\def\csname PYGdefault@tok@gh\endcsname{\let\PYGdefault@bf=\textbf\def\PYGdefault@tc##1{\textcolor[rgb]{0.00,0.00,0.50}{##1}}}
\expandafter\def\csname PYGdefault@tok@gu\endcsname{\let\PYGdefault@bf=\textbf\def\PYGdefault@tc##1{\textcolor[rgb]{0.50,0.00,0.50}{##1}}}
\expandafter\def\csname PYGdefault@tok@gd\endcsname{\def\PYGdefault@tc##1{\textcolor[rgb]{0.63,0.00,0.00}{##1}}}
\expandafter\def\csname PYGdefault@tok@gi\endcsname{\def\PYGdefault@tc##1{\textcolor[rgb]{0.00,0.63,0.00}{##1}}}
\expandafter\def\csname PYGdefault@tok@gr\endcsname{\def\PYGdefault@tc##1{\textcolor[rgb]{1.00,0.00,0.00}{##1}}}
\expandafter\def\csname PYGdefault@tok@ge\endcsname{\let\PYGdefault@it=\textit}
\expandafter\def\csname PYGdefault@tok@gs\endcsname{\let\PYGdefault@bf=\textbf}
\expandafter\def\csname PYGdefault@tok@gp\endcsname{\let\PYGdefault@bf=\textbf\def\PYGdefault@tc##1{\textcolor[rgb]{0.00,0.00,0.50}{##1}}}
\expandafter\def\csname PYGdefault@tok@go\endcsname{\def\PYGdefault@tc##1{\textcolor[rgb]{0.53,0.53,0.53}{##1}}}
\expandafter\def\csname PYGdefault@tok@gt\endcsname{\def\PYGdefault@tc##1{\textcolor[rgb]{0.00,0.27,0.87}{##1}}}
\expandafter\def\csname PYGdefault@tok@err\endcsname{\def\PYGdefault@bc##1{\setlength{\fboxsep}{0pt}\fcolorbox[rgb]{1.00,0.00,0.00}{1,1,1}{\strut ##1}}}
\expandafter\def\csname PYGdefault@tok@kc\endcsname{\let\PYGdefault@bf=\textbf\def\PYGdefault@tc##1{\textcolor[rgb]{0.00,0.50,0.00}{##1}}}
\expandafter\def\csname PYGdefault@tok@kd\endcsname{\let\PYGdefault@bf=\textbf\def\PYGdefault@tc##1{\textcolor[rgb]{0.00,0.50,0.00}{##1}}}
\expandafter\def\csname PYGdefault@tok@kn\endcsname{\let\PYGdefault@bf=\textbf\def\PYGdefault@tc##1{\textcolor[rgb]{0.00,0.50,0.00}{##1}}}
\expandafter\def\csname PYGdefault@tok@kr\endcsname{\let\PYGdefault@bf=\textbf\def\PYGdefault@tc##1{\textcolor[rgb]{0.00,0.50,0.00}{##1}}}
\expandafter\def\csname PYGdefault@tok@bp\endcsname{\def\PYGdefault@tc##1{\textcolor[rgb]{0.00,0.50,0.00}{##1}}}
\expandafter\def\csname PYGdefault@tok@fm\endcsname{\def\PYGdefault@tc##1{\textcolor[rgb]{0.00,0.00,1.00}{##1}}}
\expandafter\def\csname PYGdefault@tok@vc\endcsname{\def\PYGdefault@tc##1{\textcolor[rgb]{0.10,0.09,0.49}{##1}}}
\expandafter\def\csname PYGdefault@tok@vg\endcsname{\def\PYGdefault@tc##1{\textcolor[rgb]{0.10,0.09,0.49}{##1}}}
\expandafter\def\csname PYGdefault@tok@vi\endcsname{\def\PYGdefault@tc##1{\textcolor[rgb]{0.10,0.09,0.49}{##1}}}
\expandafter\def\csname PYGdefault@tok@vm\endcsname{\def\PYGdefault@tc##1{\textcolor[rgb]{0.10,0.09,0.49}{##1}}}
\expandafter\def\csname PYGdefault@tok@sa\endcsname{\def\PYGdefault@tc##1{\textcolor[rgb]{0.73,0.13,0.13}{##1}}}
\expandafter\def\csname PYGdefault@tok@sb\endcsname{\def\PYGdefault@tc##1{\textcolor[rgb]{0.73,0.13,0.13}{##1}}}
\expandafter\def\csname PYGdefault@tok@sc\endcsname{\def\PYGdefault@tc##1{\textcolor[rgb]{0.73,0.13,0.13}{##1}}}
\expandafter\def\csname PYGdefault@tok@dl\endcsname{\def\PYGdefault@tc##1{\textcolor[rgb]{0.73,0.13,0.13}{##1}}}
\expandafter\def\csname PYGdefault@tok@s2\endcsname{\def\PYGdefault@tc##1{\textcolor[rgb]{0.73,0.13,0.13}{##1}}}
\expandafter\def\csname PYGdefault@tok@sh\endcsname{\def\PYGdefault@tc##1{\textcolor[rgb]{0.73,0.13,0.13}{##1}}}
\expandafter\def\csname PYGdefault@tok@s1\endcsname{\def\PYGdefault@tc##1{\textcolor[rgb]{0.73,0.13,0.13}{##1}}}
\expandafter\def\csname PYGdefault@tok@mb\endcsname{\def\PYGdefault@tc##1{\textcolor[rgb]{0.40,0.40,0.40}{##1}}}
\expandafter\def\csname PYGdefault@tok@mf\endcsname{\def\PYGdefault@tc##1{\textcolor[rgb]{0.40,0.40,0.40}{##1}}}
\expandafter\def\csname PYGdefault@tok@mh\endcsname{\def\PYGdefault@tc##1{\textcolor[rgb]{0.40,0.40,0.40}{##1}}}
\expandafter\def\csname PYGdefault@tok@mi\endcsname{\def\PYGdefault@tc##1{\textcolor[rgb]{0.40,0.40,0.40}{##1}}}
\expandafter\def\csname PYGdefault@tok@il\endcsname{\def\PYGdefault@tc##1{\textcolor[rgb]{0.40,0.40,0.40}{##1}}}
\expandafter\def\csname PYGdefault@tok@mo\endcsname{\def\PYGdefault@tc##1{\textcolor[rgb]{0.40,0.40,0.40}{##1}}}
\expandafter\def\csname PYGdefault@tok@ch\endcsname{\let\PYGdefault@it=\textit\def\PYGdefault@tc##1{\textcolor[rgb]{0.25,0.50,0.50}{##1}}}
\expandafter\def\csname PYGdefault@tok@cm\endcsname{\let\PYGdefault@it=\textit\def\PYGdefault@tc##1{\textcolor[rgb]{0.25,0.50,0.50}{##1}}}
\expandafter\def\csname PYGdefault@tok@cpf\endcsname{\let\PYGdefault@it=\textit\def\PYGdefault@tc##1{\textcolor[rgb]{0.25,0.50,0.50}{##1}}}
\expandafter\def\csname PYGdefault@tok@c1\endcsname{\let\PYGdefault@it=\textit\def\PYGdefault@tc##1{\textcolor[rgb]{0.25,0.50,0.50}{##1}}}
\expandafter\def\csname PYGdefault@tok@cs\endcsname{\let\PYGdefault@it=\textit\def\PYGdefault@tc##1{\textcolor[rgb]{0.25,0.50,0.50}{##1}}}

\makeatother

\begin{document}

\maketitle
\thispagestyle{empty}
\pagestyle{empty}

\begin{abstract}
  Invasive cardiac catheterisation is a common procedure that is carried
  out before surgical intervention. Yet, invasive cardiac diagnostics
  are full of risks, especially for young children. Decades of research
  has been conducted on the so called \textit{inverse problem of
    electrocardiography}, which can be used to reconstruct \textit{Heart
    Surface Potentials} (HSPs) from \textit{Body Surface Potentials}
  (BSPs), for non-invasive diagnostics. State of the art solutions to
  the inverse problem are unsatisfactory, since the inverse problem is
  known to be \textit{ill-posed}. In this paper we propose a novel
  approach to reconstructing HSPs from BSPs using a \textit{Time-Delay
    Artificial Neural Network} (TDANN). We first design the TDANN
  architecture, and then develop an iterative search space algorithm to
  find the parameters of the TDANN, which results in the best overall
  HSP prediction. We use real-world recorded BSPs and HSPs from
  individuals suffering from serious cardiac conditions to validate our
  TDANN. The results are encouraging, in that coefficients obtained by
  correlating the predicted HSP with the recorded patient' HSP approach
  ideal values.
\end{abstract}

\section{Introduction and Related Work}
\label{sec:intr-relat-work}

Cardiac catheterisation is an essential procedure carried out to
diagnose heart tissue ailments before surgical interventions.
Catheterisation is associated with risks such as stroke, heart attacks,
and even death~\cite{al2017safety}, especially among young
children~\cite{kreutzer2015catastrophic,o2015predictors}. Unlike
catheterisation, \textit{Body Surface Potentials} (BSP), captured
non-invasively, can potentially be used for diagnosis. The primary idea
is to \textit{reconstruct} the \textit{Heart Surface Potentials}, (HSP),
used for diagnosis, from the BSPs.

The traditional approach to solving this reconstruction problem is to
first model the human torso as a series of \textit{Partial Differential
  Equations} (PDE), which relates the BSPs and HSPs and then inverting
this torso model to produce the HSPs from BSPs. Finding a solution to
the inverted torso model is termed the \textit{inverse problem of
  electrocardiography}~\cite{rudy1988inverse}. The inverse problem of
electrocardiography is mathematically
proven~\cite{rudy1988inverse,hullerum2012inverse} to be
\textit{ill-posed}. Thus, small changes in the BSP inputs can lead to
distortions and unbounded errors in the HSP
outputs~\cite{hullerum2012inverse}.

Given $N$ BSPs, represented by the vector $\Phi_{B}$ and $M$ HSPs,
represented by vector $\Phi_{H}$. The inverse problem is given by the
linear matrix equation \mbox{$\Phi_{B} = T_{BH} \times \Phi_{H}$}. A
number of techniques from standard numerical analysis have been proposed
to approximate solutions to the transfer matrix $T_{BH}$. A good
overview is provided in~\cite{gulrajani1998forward1}. The primary idea
in all proposed techniques and their variants is to select elements of
the matrix $T_{BH}$ such that the error
\mbox{$\|\Phi_{B} - T_{BH} \times \Phi_{H}\|^{2}$}~\footnote{$\|.\|$ is
  the l2-norm.}, is minimized.

The very large search space of the aforementioned minimization problem
is reduced by the technique called Tikhonov
regularization~\cite{tikhonov1977solutions}, where the minimization
objective is transformed to:
$\| \Phi_{B} - T_{BH} \times \Phi_{H} \|^{2} + \gamma \times \|
\mathbf{C} \times \Phi_{H}\|^{2}$, where $M \times M$ matrix
$\mathbf{C}$ is the discrete approximation of the surface laplacian, and
$\gamma$ is the regularization parameter, which controls the weight of
the constraint condition. As $\gamma$ approaches zero, the solution
oscillates, due to the ill-posed nature of the problem. On the other
hand, a large value of $\gamma$ leads to overly smooth solution,
unrepresentative of the real HSPs. Hence, the numerical solutions to the
regularization problem give unsatisfactory
results~\cite{ghosh2009application}.

Recently, machine learning solutions have been proposed to reconstruct
HSPs from BSPs. In~\cite{zemzemi2013machine}, kernel ridge regression,
with a Gaussian kernel, is used to reconstruct HSPs from BSPs. Although
performing better than the numerical solutions, the machine learning
solution has low correlation with the exact solution. Hence, the
activation map of the heart surface potentials is unsatisfactory.
Furthermore, none of the techniques consider reconstruction of HSPs
under cardiac disease states, therefore, the fidelity of reconstructed
activation map, under diseased states is suspect.

This paper has three major contributions:
\begin{enumerate}
\item We propose a neural network model that can predic HSPs from BSPs.
\item We show the efficacy of the proposed neural network model under
  normal and diseased heart conditions, in particular, for patients
  suffering from ventricular flutter.
\item We quantify the validity of the HSP activation maps, by
  correlating them to real-world recorded patient data.
\end{enumerate}

 \section{A time-delay artificial neural network for predicting heart
  surface potentials}
\label{sec:preliminaries}

Our objective is to develop an \textit{Artificial Neural Network} (ANN)
architecture that predicts a single HSP from a single BSP. In this
section, we first describe our ANN architecture and then propose search
space algorithm to predict parameters of the proposed ANN.

\subsection{The basic artificial neural network architecture}
\label{sec:artif-neur-netw}

\begin{figure*}[tb]
  \centering

  \subfloat[A standard artificial neural
  network~\label{fig:1a}]{\includegraphics[scale=0.4]{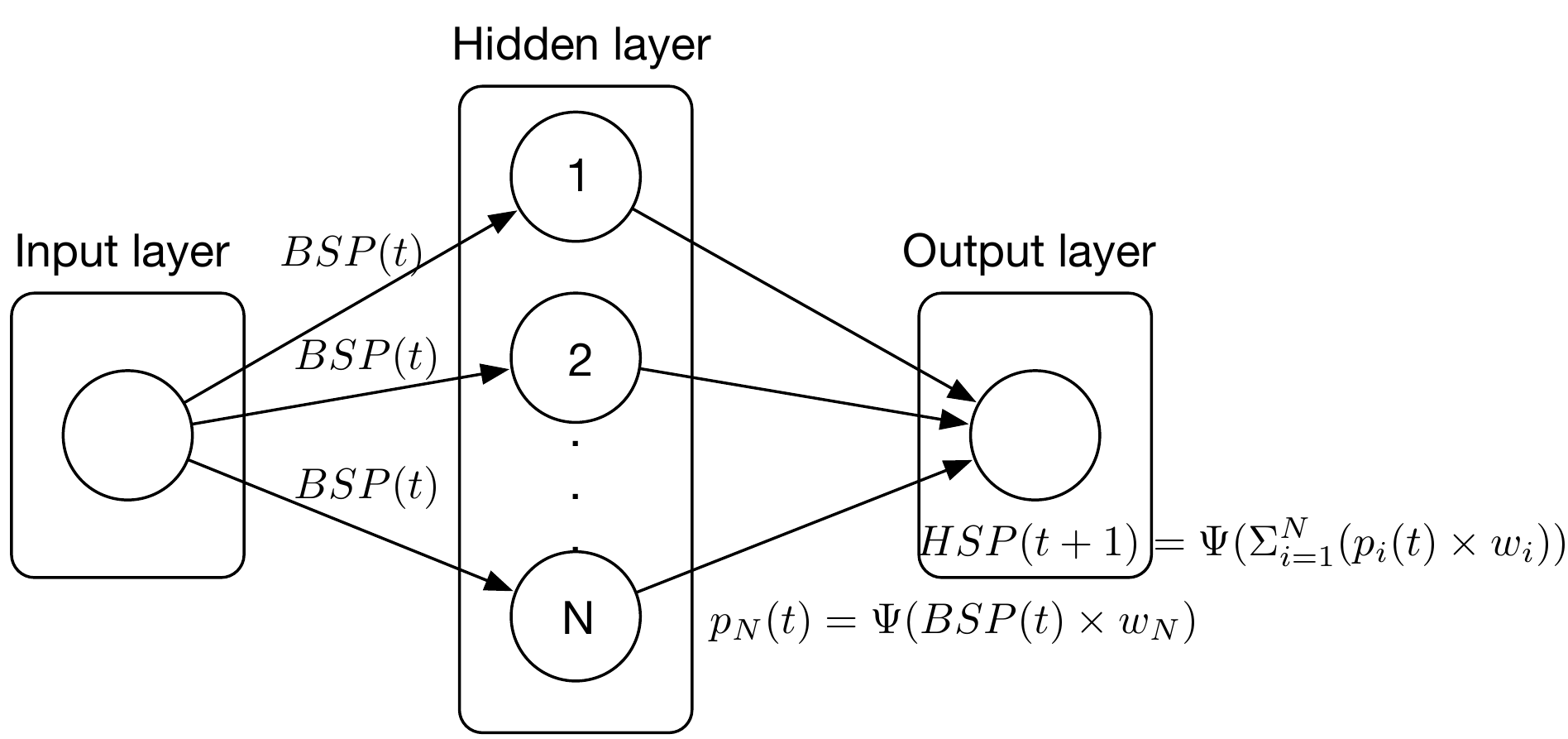}}
  \subfloat[A time-delayed
  neuron~\label{fig:1b}]{\includegraphics[scale=0.45]{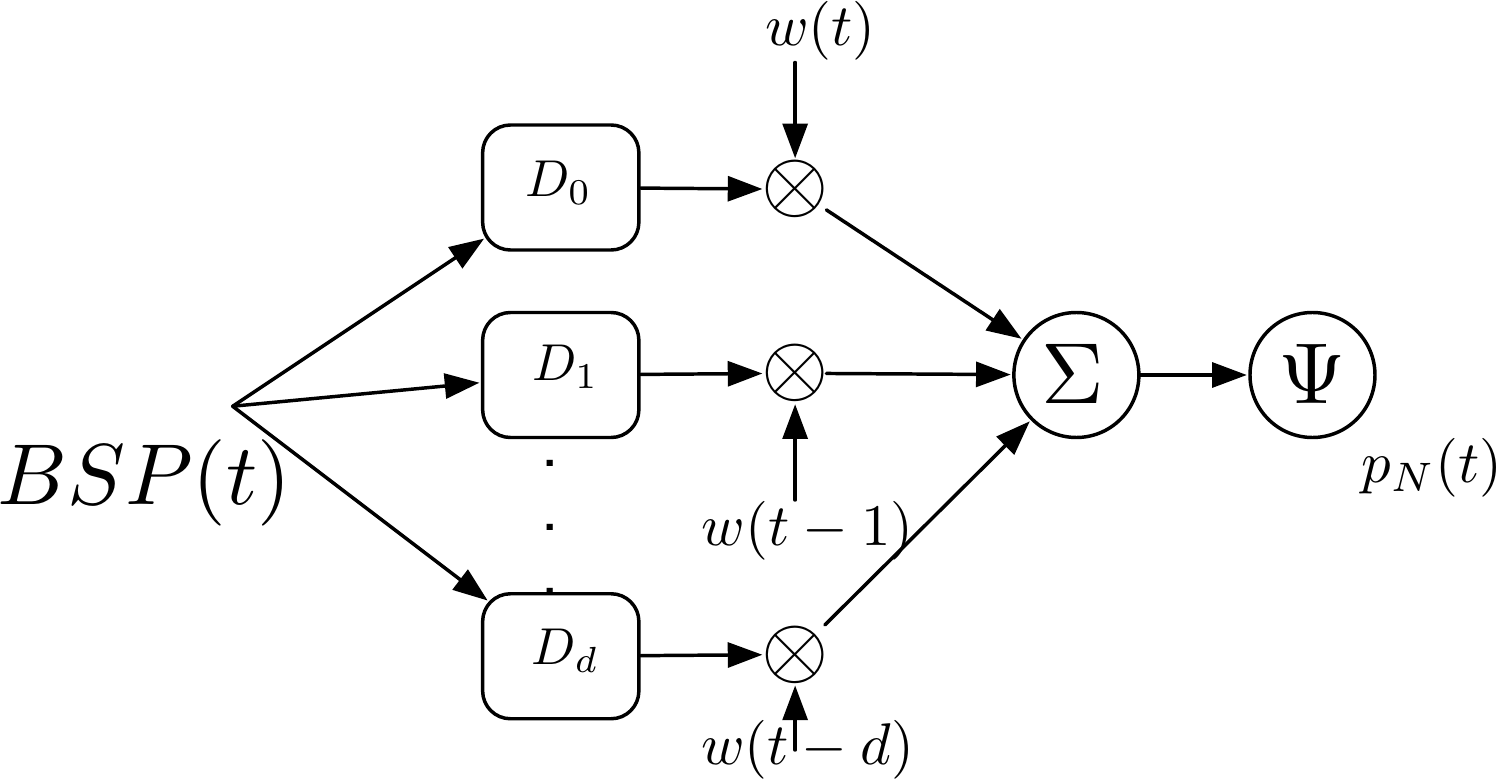}}

  \caption{The artificial neural network architecture and a time-delayed
    neuron in the hidden layer}
  \label{fig:1}
\end{figure*}

The ANN architecture that forms the basis for predicting the HSP from
the BSP is shown in Figure~\ref{fig:1a}. The ANN consists of three major
components: \textcircled{1} an input layer that reads the input
signal(s). In our case the input layer consists of a single neuron
(shown as a circle), which reads the BSP time series. \textcircled{2} A
hidden layer, which consists of $N$ neurons, which \textit{learn} the
relationship between the BSP and the HSP. \textcircled{3} An output
layer, which produces the HSP. In our case the output layer consists of
a single neuron, because we produce a single HSP time series. A so
called \textit{weight} (vector), e.g., $w_{N}$ in Figure~\ref{fig:1a},
and an \textit{activation function} $\Psi$ are associated with each
neuron in the network. The output of a neuron is the inner product of
the input vector and the weight vector applied to the activation
function.

The number of neurons in the hidden layer, the weight vector associated
with each neuron, and the activation function together decide the
efficacy of the ANN. Neural network weights are chosen to minimize the
mean square error using Levenberg~\cite{more1978levenberg} algorithm.
Furthermore, the activation function is the sigmoid function. The number
of neurons in the hidden layer are chosen using cross-validation
techniques described later in Section~\ref{sec:train-algor-artif}.

\subsection{Time-delayed artificial neuron}
\label{sec:time-delay-artif}

Each neuron uses the current and $d$ previous values of the input BSP to
predict the output HSP, at any given point in time, as shown in
Figure~\ref{fig:1b}. Such an ANN architecture is called a
\textit{Time-Delayed Artificial Neural Network}
(TDANN)~\cite{waibel1989phoneme}. Hence, for some neuron $N$, in the
hidden layer, the output is given by:
\mbox{$p_{N}(t) = \Psi(\Sigma_{j=0}^{d}BSP(t-j)\times w(t-j))$}. We use
a TDANN, because there is a very high correlation between any point in
the BSP(t) and its previous values BSP(t-1), BSP(t-2), $\ldots$,
BSP(t-d), since heart has a rhythm under normal and diseased states.

\subsection{Training the time-delayed artificial neural network}
\label{sec:train-algor-artif}

There are three parameters that need to be tuned in our TDANN:
\textcircled{1} the number of neurons $N$ in the hidden layer,
\textcircled{2} the weight vector of each neuron in the network, and
\textcircled{3} the time-delay window $d$ for each neuron in the hidden
layer. We perform exhaustive search space exploration, using iterative
approach to tune the three parameters. The search space exploration
algorithm, in Matlab, is given in Figure~\ref{fig:iteralgo}.

Given $P$ patient data, we use $P-1$ BSP and HSP time series for
training the TDANN and one patient recording footage for testing
purposes. In Figure~\ref{fig:iteralgo}, BSP1, BSP2, HSP1, and HSP2,
represent the training input and output datasets, respectively. BSP3 and
HSP3 are the test datasets. First the training input and output sets are
concatenated together into a single time series vector
(lines~\ref{lst:1} and~\ref{lst:2}). Next, the maximum delay window size
$d=20$ is initialized. This value is obtained using autocorrelation of
BSP training time series. Next, a vector of neurons in the hidden layer
is initialized (line~\ref{lst:3}). In this case, a row vector of 1-20
neurons and then 40, 80, and 100 neurons will be explored during the
execution of the iterative algorithm. Line~\ref{lst:4} initializes the
variable that will hold the best Pearson correlation coefficient for the
test dataset during search space exploration.

The actual iteration, selecting the best number of neurons in the hidden
layer and the time delay window happens from
lines~\ref{lst:5}-\ref{lst:6}. The algorithm iterates through each
neuron in the \texttt{neurons} vector, and for each hidden layer size,
iterates through each possible delay window size. Line~\ref{lst:7}
initializes the TDANN. Line~\ref{lst:8} trains the network over the
training dataset. During training, the weight vector for each neuron is
obtained using Levenberg minimization
algorithm~\cite{more1978levenberg}. Next, the algorithm uses the
\textit{never seen before} test dataset BSP3 to predict the heart
surface potential output (\texttt{PHSP3}) and correlates it to the
actual HSP3 dataset. The net, along with its configuration, that gives
the best correlation coefficient is stored for later use
(lines~\ref{lst:9}-line~\ref{lst:10}).

\newsavebox{\overallsmt}
\begin{lrbox}{\overallsmt}
  \begin{scriptsize}
  \begin{minipage}{\linewidth}
    \begin{Verbatim}[commandchars=\\\{\},codes={\catcode`\$=3\catcode`\^=7\catcode`\_=8}]
\PYG{n}{clear}\PYG{p}{;} \PYG{n}{clc}\PYG{p}{;}
\PYG{c}{\PYGZpc{}\PYGZpc{} Training on patient\PYGZhy{}1 and patient\PYGZhy{}2 together}
\PYG{n}{train\PYGZus{}input} \PYG{p}{=} \PYG{n}{dtrend}\PYG{p}{(}\PYG{p}{[}\PYG{n}{BSP1}\PYG{p}{(}\PYG{n}{s}\PYG{p}{:}\PYG{n}{e}\PYG{p}{)}\PYG{p}{;} \PYG{n}{BSP2}\PYG{p}{(}\PYG{n}{s}\PYG{p}{:}\PYG{n}{e}\PYG{p}{)}\PYG{p}{]}\PYG{p}{)}\PYG{p}{;} \PYG{esc}{\label{lst:1}}
\PYG{n}{train\PYGZus{}output} \PYG{p}{=} \PYG{n}{dtrend}\PYG{p}{(}\PYG{p}{[}\PYG{n}{HSP1}\PYG{p}{(}\PYG{n}{s}\PYG{p}{:}\PYG{n}{e}\PYG{p}{)}\PYG{p}{;} \PYG{n}{HSP2}\PYG{p}{(}\PYG{n}{s}\PYG{p}{:}\PYG{n}{e}\PYG{p}{)}\PYG{p}{]}\PYG{p}{)}\PYG{p}{;} \PYG{esc}{\label{lst:2}}

\PYG{n}{d} \PYG{p}{=} \PYG{l+m+mi}{20}\PYG{p}{;} \PYG{c}{\PYGZpc{}\PYGZpc{} The maximum delay}
\PYG{c}{\PYGZpc{}\PYGZpc{} The neuron vector}
\PYG{n}{neurons} \PYG{p}{=} \PYG{p}{[}\PYG{l+m+mi}{1}\PYG{p}{:}\PYG{l+m+mi}{1}\PYG{p}{:}\PYG{l+m+mi}{20}\PYG{p}{,} \PYG{l+m+mi}{40}\PYG{p}{,} \PYG{l+m+mi}{80}\PYG{p}{,} \PYG{l+m+mi}{100}\PYG{p}{]}\PYG{p}{;} \PYG{esc}{\label{lst:3}}

\PYG{n}{maxC} \PYG{p}{=} \PYG{l+m+mf}{0.0} \PYG{esc}{\label{lst:4}}

\PYG{k}{for} \PYG{n}{n} \PYG{p}{=} \PYG{l+m+mi}{1}\PYG{p}{:}\PYG{n+nb}{size}\PYG{p}{(}\PYG{n}{neurons}\PYG{p}{)}\PYG{p}{(}\PYG{l+m+mi}{2}\PYG{p}{)}  \PYG{esc}{\label{lst:5}} \PYG{c}{\PYGZpc{} Iterate through neurons}
 \PYG{k}{for} \PYG{n}{delay} \PYG{p}{=} \PYG{l+m+mi}{1}\PYG{p}{:}\PYG{n}{d} \PYG{c}{\PYGZpc{} Iterate through delay}
  \PYG{p}{[}\PYG{n}{X}\PYG{p}{,} \PYG{o}{\PYGZti{}}\PYG{p}{]} \PYG{p}{=} \PYG{n}{tonndata}\PYG{p}{(}\PYG{n}{train\PYGZus{}input}\PYG{p}{,} \PYG{n}{false}\PYG{p}{,} \PYG{n}{false}\PYG{p}{)}\PYG{p}{;}
  \PYG{p}{[}\PYG{n}{T}\PYG{p}{,} \PYG{o}{\PYGZti{}}\PYG{p}{]} \PYG{p}{=} \PYG{n}{tonndata}\PYG{p}{(}\PYG{n}{train\PYGZus{}output}\PYG{p}{,} \PYG{n}{false}\PYG{p}{,} \PYG{n}{false}\PYG{p}{)}\PYG{p}{;}
  \PYG{c}{\PYGZpc{}\PYGZpc{} Initialize a time\PYGZhy{}delay network}
  \PYG{n}{net} \PYG{p}{=} \PYG{n}{timedelaynet}\PYG{p}{(}\PYG{p}{(}\PYG{l+m+mi}{0}\PYG{p}{:}\PYG{n}{delay}\PYG{p}{)}\PYG{p}{,} \PYG{n}{neurons}\PYG{p}{(}\PYG{n}{n}\PYG{p}{)}\PYG{p}{,}
                     \PYG{l+s}{\PYGZsq{}}\PYG{l+s}{trainlm\PYGZsq{}}\PYG{p}{)}\PYG{p}{;} \PYG{esc}{\label{lst:7}}
  \PYG{c}{\PYGZpc{}\PYGZpc{} Train the network.}
  \PYG{c}{\PYGZpc{}\PYGZpc{} Selecting weights via levenberg algorithm}
  \PYG{p}{[}\PYG{n}{Xs}\PYG{p}{,}\PYG{n}{Xi}\PYG{p}{,}\PYG{n}{Ai}\PYG{p}{,}\PYG{n}{Ts}\PYG{p}{]} \PYG{p}{=} \PYG{n}{preparets}\PYG{p}{(}\PYG{n}{net}\PYG{p}{,}\PYG{n}{X}\PYG{p}{,}\PYG{n}{T}\PYG{p}{)}\PYG{p}{;}
  \PYG{n}{net} \PYG{p}{=} \PYG{n}{train}\PYG{p}{(}\PYG{n}{net}\PYG{p}{,}\PYG{n}{Xs}\PYG{p}{,}\PYG{n}{Ts}\PYG{p}{,} \PYG{n}{Xi}\PYG{p}{,} \PYG{n}{Ai}\PYG{p}{)}\PYG{p}{;} \PYG{esc}{\label{lst:8}}
  \PYG{c}{\PYGZpc{} compute correlation coefficient}
  \PYG{c}{\PYGZpc{} for never seen before test\PYGZhy{}patient.}

  \PYG{c}{\PYGZpc{} PHSP3 is the predicted output for BSP3}
  \PYG{n}{PHSP3} \PYG{p}{=} \PYG{n}{net}\PYG{p}{(}\PYG{n}{BSP3}\PYG{p}{(}\PYG{n}{s}\PYG{p}{:}\PYG{n}{e}\PYG{p}{)}\PYG{p}{,} \PYG{n}{false}\PYG{p}{,} \PYG{n}{false}\PYG{p}{,}
             \PYG{n}{BSP3}\PYG{p}{(}\PYG{n}{s}\PYG{p}{:}\PYG{n}{s}\PYG{o}{+}\PYG{p}{(}\PYG{n}{delay}\PYG{o}{\PYGZhy{}}\PYG{l+m+mi}{1}\PYG{p}{)}\PYG{p}{)}\PYG{p}{,} \PYG{n}{false}\PYG{p}{,} \PYG{n}{false}\PYG{p}{)}\PYG{p}{;} \PYG{esc}{\label{lst:9}}
  \PYG{n}{C} \PYG{p}{=} \PYG{n}{corr}\PYG{p}{(}\PYG{n}{HSP3}\PYG{p}{,} \PYG{n}{PHSP3}\PYG{p}{(}\PYG{n}{s}\PYG{p}{:}\PYG{n}{e}\PYG{p}{)}\PYG{p}{)}\PYG{p}{;} \PYG{c}{\PYGZpc{} pearson correlation}
  \PYG{k}{if} \PYG{n}{C} \PYG{o}{\PYGZgt{}} \PYG{n}{maxC}
   \PYG{n}{maxC} \PYG{p}{=} \PYG{n}{C}\PYG{p}{;}
   \PYG{c}{\PYGZpc{}\PYGZpc{} Store result}
   \PYG{n}{result} \PYG{p}{=} \PYG{p}{[}\PYG{n}{net}\PYG{p}{,} \PYG{n}{maxC}\PYG{p}{,} \PYG{n}{neurons}\PYG{p}{(}\PYG{n}{n}\PYG{p}{)}\PYG{p}{,} \PYG{n}{delay}\PYG{p}{]}\PYG{p}{;} \PYG{esc}{\label{lst:10}}
  \PYG{k}{end}
 \PYG{k}{end}
\PYG{k}{end} \PYG{esc}{\label{lst:6}}
\end{Verbatim}





  \end{minipage}
  \end{scriptsize}
\end{lrbox}

\begin{figure}[th]
  \centering
  {\scalebox{0.8}{\usebox{\overallsmt}}}
  \caption{The algorithm to tune TDANN parameters}
  \label{fig:iteralgo}
\end{figure}

\section{Experimental Benchmarking}
\label{sec:experimental-results}

In this section we describe the efficacy of the proposed TDANN using
real-world patient recordings.

\subsection{Experimental setup}
\label{sec:experimental-setup}

We use the BSP and HSP data recorded from patients, available
in~\cite{annaael}. We use the BSP recorded from Lead I of the
electrocardiogram machine as the input and the unipolar HSP recorded, at
the same time, from the Right-Ventricular Apex, as the output, for all
our experiments. The TDANN training and testing algorithm is designed in
Matlab version R2016b, running on OSX 10.11.6, Intel Core i5 2.9 GHz
laptop with 8 GB of RAM.

\begin{table}[tb]
  \centering
    \begin{tabular}{|c|c|c|c|p{30pt}|}
      \hline
      Patient ID & Heart rhythm& Start (sec) & End (sec) & Used for \\
      \hline
      343220 & Normal rhythm & 1 & 10000 & Training\\
      \hline
      33093 & Normal rhythm & 1 & 10000 & Training\\
      \hline
      343220 & Normal rhythm & 3000 & 65000 & Validation\\
      \hline
      33093 & Normal rhythm & 3000 & 65000 & Validation\\
      \hline
      221708 & Normal rhythm & 1 & 65000 & Testing\\
      \hline
      343300 & Ventricular flutter & 3000 & 6000 & Training \& Validation\\
      \hline
      176230 & Ventricular flutter & 3000 & 6000 & Training \& Validation\\
      \hline
      198a385 & Ventricular flutter & 10000 & 15000 & Testing\\
      \hline
    \end{tabular}
  \caption{Patient data used for experiments}
  \label{tab:1}
\end{table}

The patient data from~\cite{annaael} used in our experiments is
tabulated in Table~\ref{tab:1}. Column-1 lists the specific footage used
for experiments from the database. Column-2 lists the heart condition.
We trained our TDANN to predict HSP under normal heart rate and
ventricular flutter. Ventricular flutter is especially important,
because it is transient and can lead to death. Next the table lists the
starting and ending seconds of the complete footage that was used for
experiments. Finally, the table lists the reason for using each of these
footages, some were used for training, while others were used for
cross-validation and testing purposes.

\subsection{Results}
\label{sec:results}

We present two sets of results: \textcircled{1} the pearson correlation
coefficients obtained from the best trained neural nets for the
validation and testing datasets from Table~\ref{tab:1}. We use the
training algorithm described previously in Figure~\ref{fig:iteralgo}.
\textcircled{2} The affect on prediction accuracy with changing delay
window ($d$) and number of neurons ($N$) in the hidden layer. In all the
figures that follow, we plot only a part of the correlated time-series
for ease of understanding.

\begin{figure}[th]
  \centering
  {\includegraphics[scale=0.53]{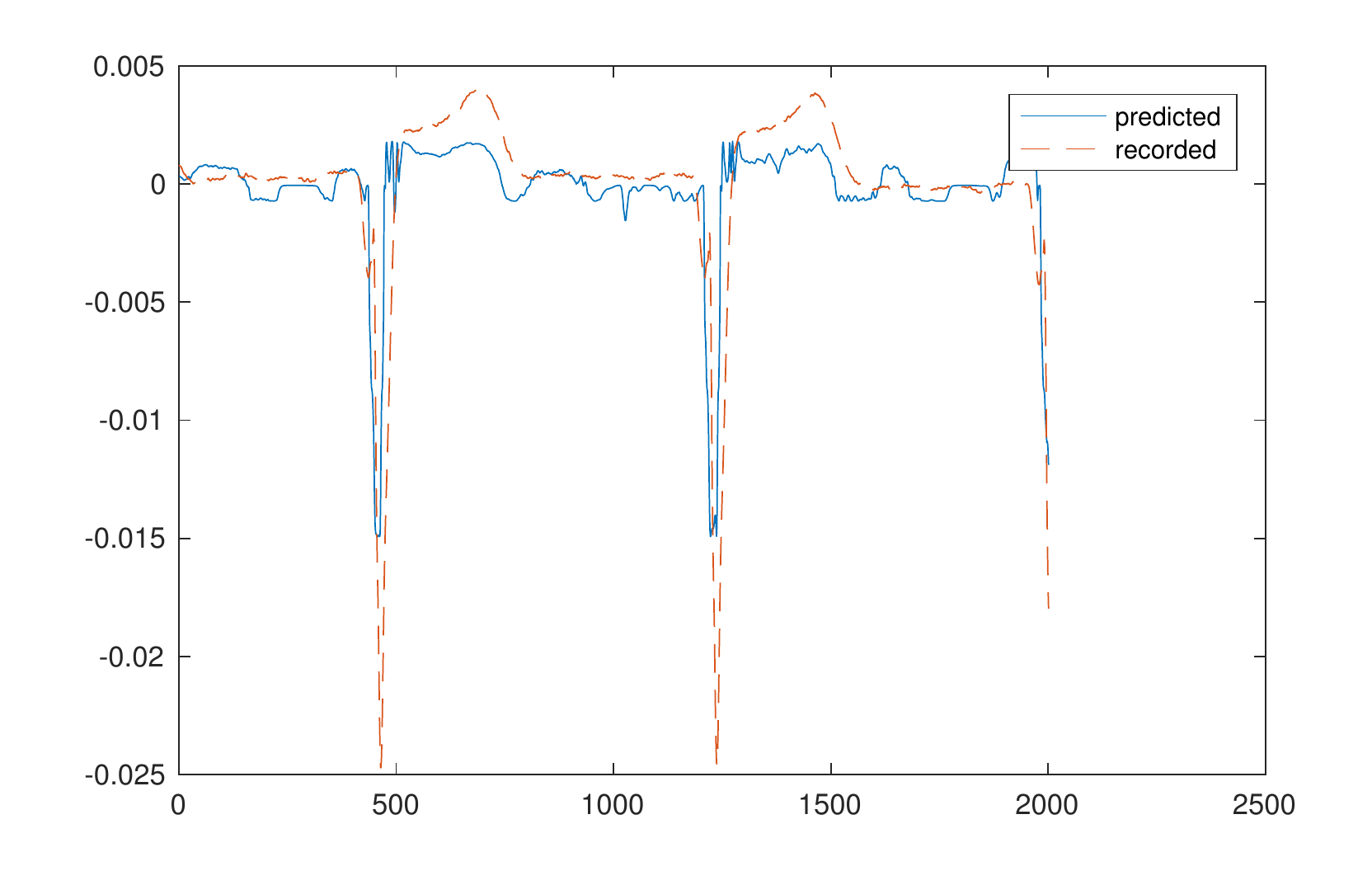}} 
  \caption{Validation result Patient: 33093, normal heart rhythm,
    correlation coefficient: 0.7, $N = 11$, $d = 15$.}
  \label{fig:res1a}
\end{figure}

\begin{figure}[th]
  \centering
  {\includegraphics[scale=0.53]{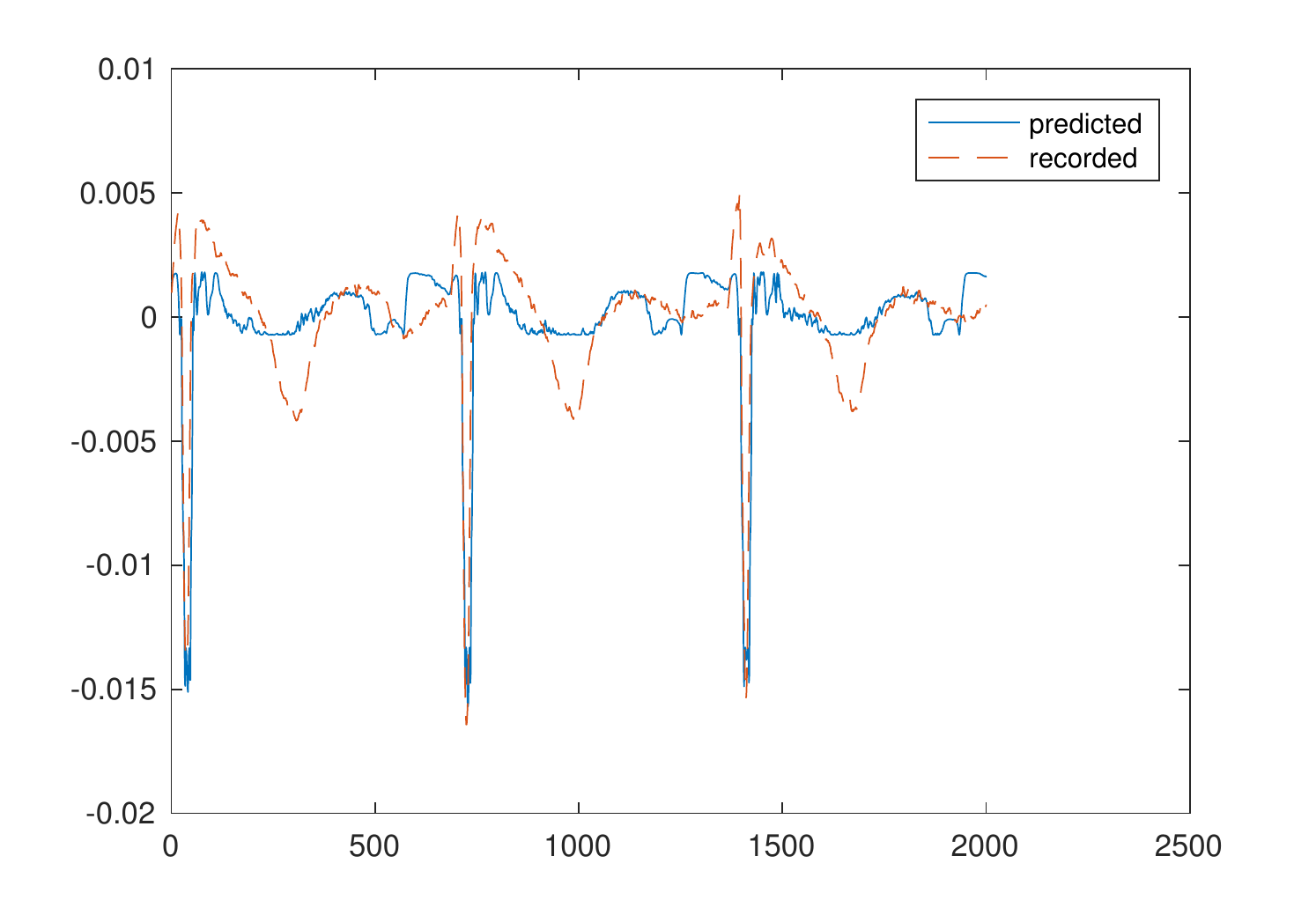}} 
  \caption{Validation result Patient: 343330, normal heart rhythm,
    correlation coefficient: 0.7519, $N=11$, $d=15$.}
  \label{fig:res1b}
\end{figure}

\begin{figure}[th]
  \centering
  {\includegraphics[scale=0.53]{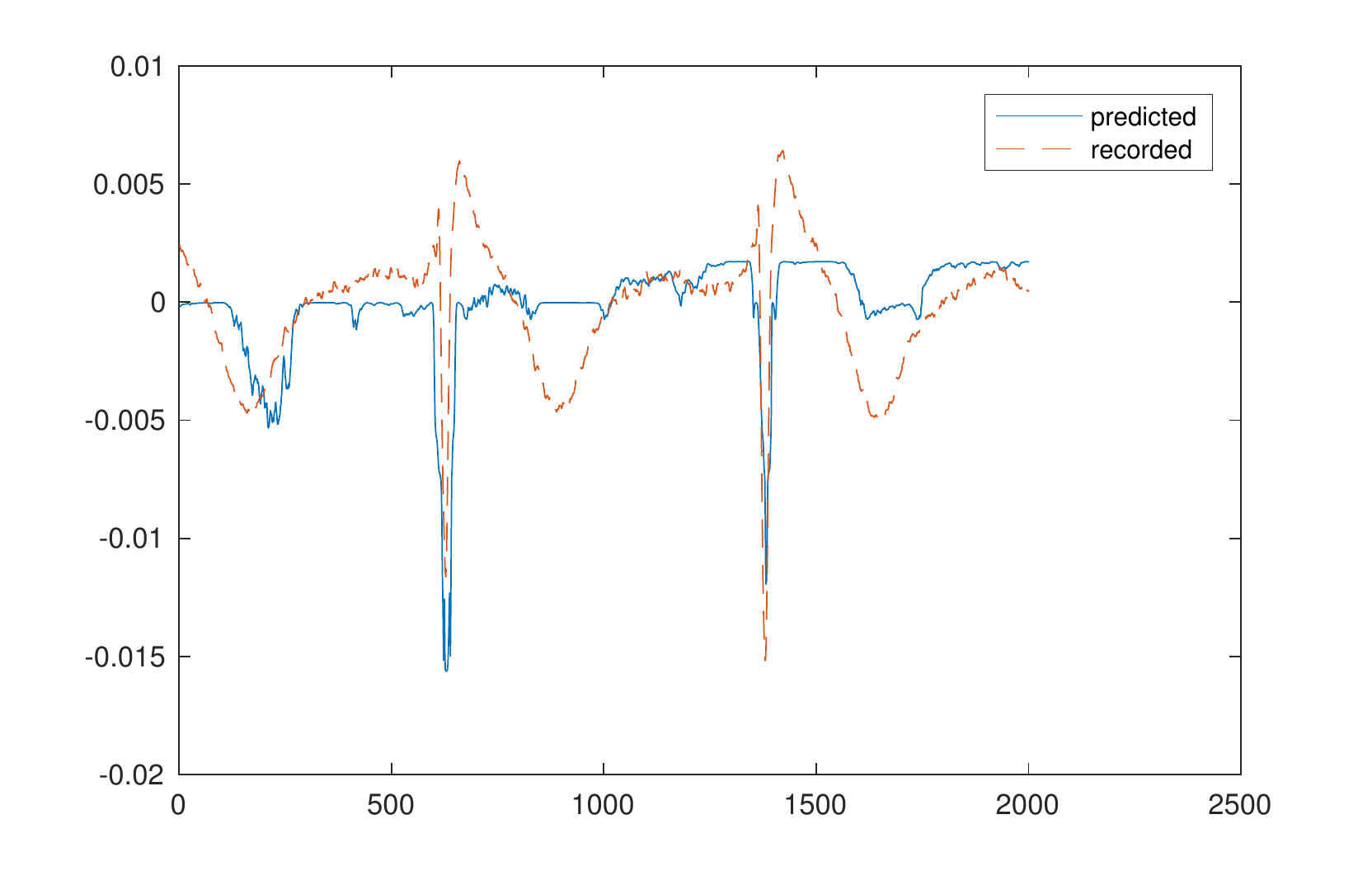}} 
  \caption{Test result Patient: 221708, normal heart rhythm,
    correlation coefficient: 0.5122, $N=11$, $d=15$.}
  \label{fig:res1c}
\end{figure}

Figures~\ref{fig:res1a}-\ref{fig:res1c} correlate the predicted HSP with
the recorded patient HSP at the right ventricular atrium under normal
heart rhythm. In all cases the results are quite consistent. Especially
note that the voltage peaks are consistently met in all cases, a problem
that is faced by the current state of the art techniques.

\begin{figure}[t]
  \centering
  {\includegraphics[scale=0.53]{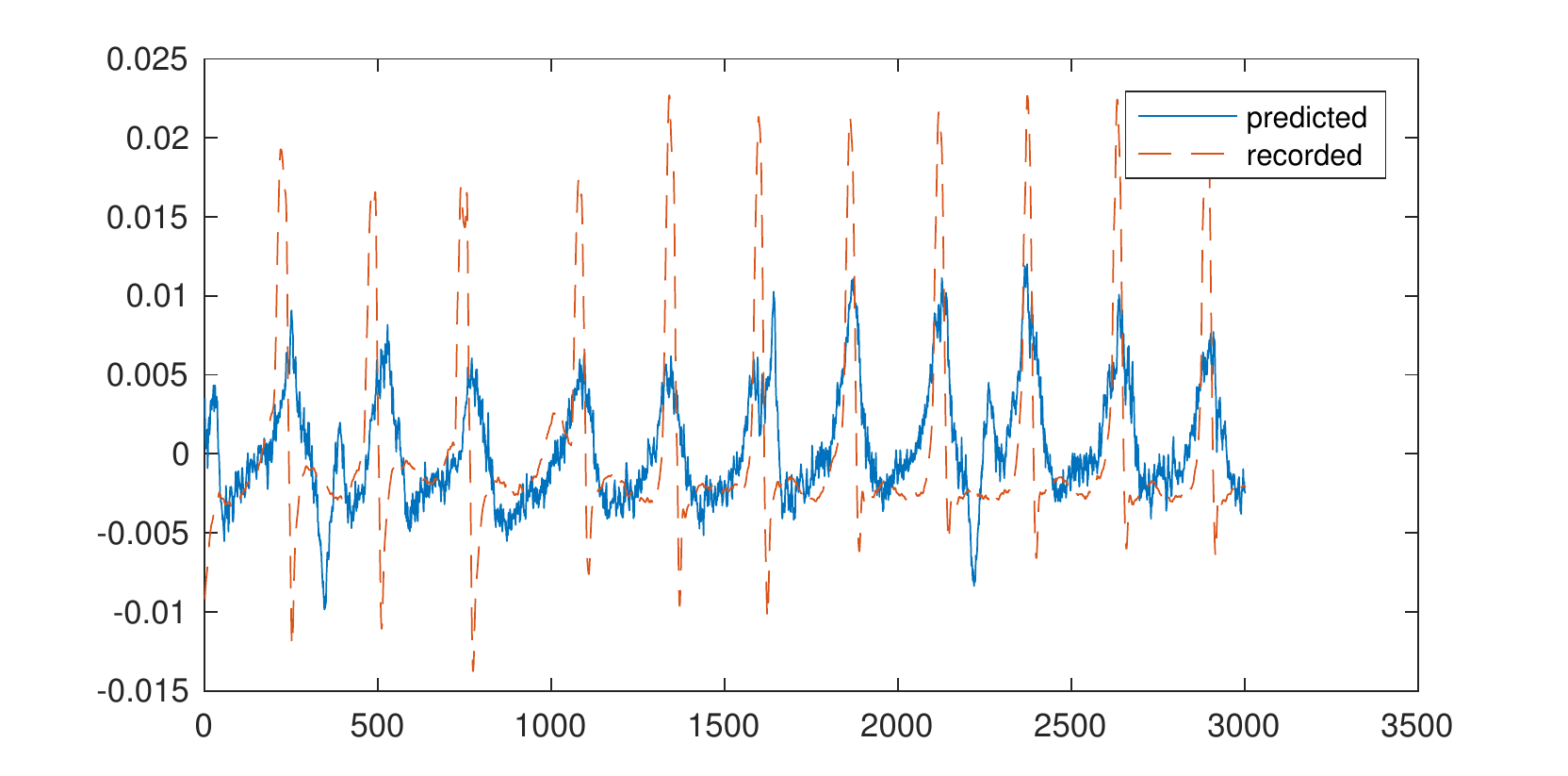}} 
  \caption{Validation result Patient: 343330, ventricular flutter,
    correlation coefficient: 0.45536, $N = 5$, $d = 18$.}
  \label{fig:res2a}
\end{figure}

\begin{figure}[t]
  \centering
  \includegraphics[scale=0.53]{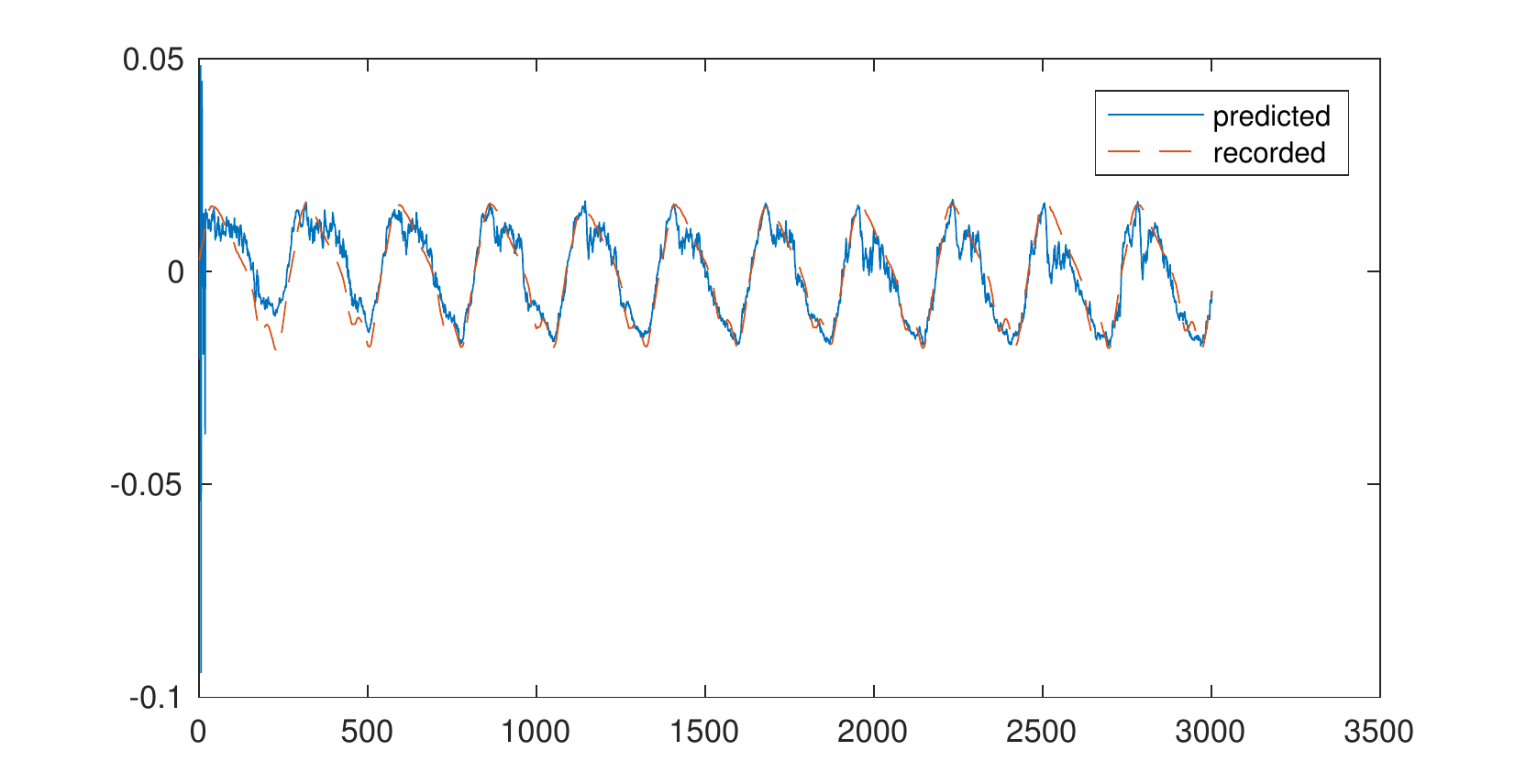} 
  \caption{Validation result Patient: 176230, ventricular flutter,
    correlation coefficient: 0.9, $N=5$, $d=18$.}
  \label{fig:res2b}
\end{figure}

\begin{figure}[h]
  \centering
  {\includegraphics[scale=0.53]{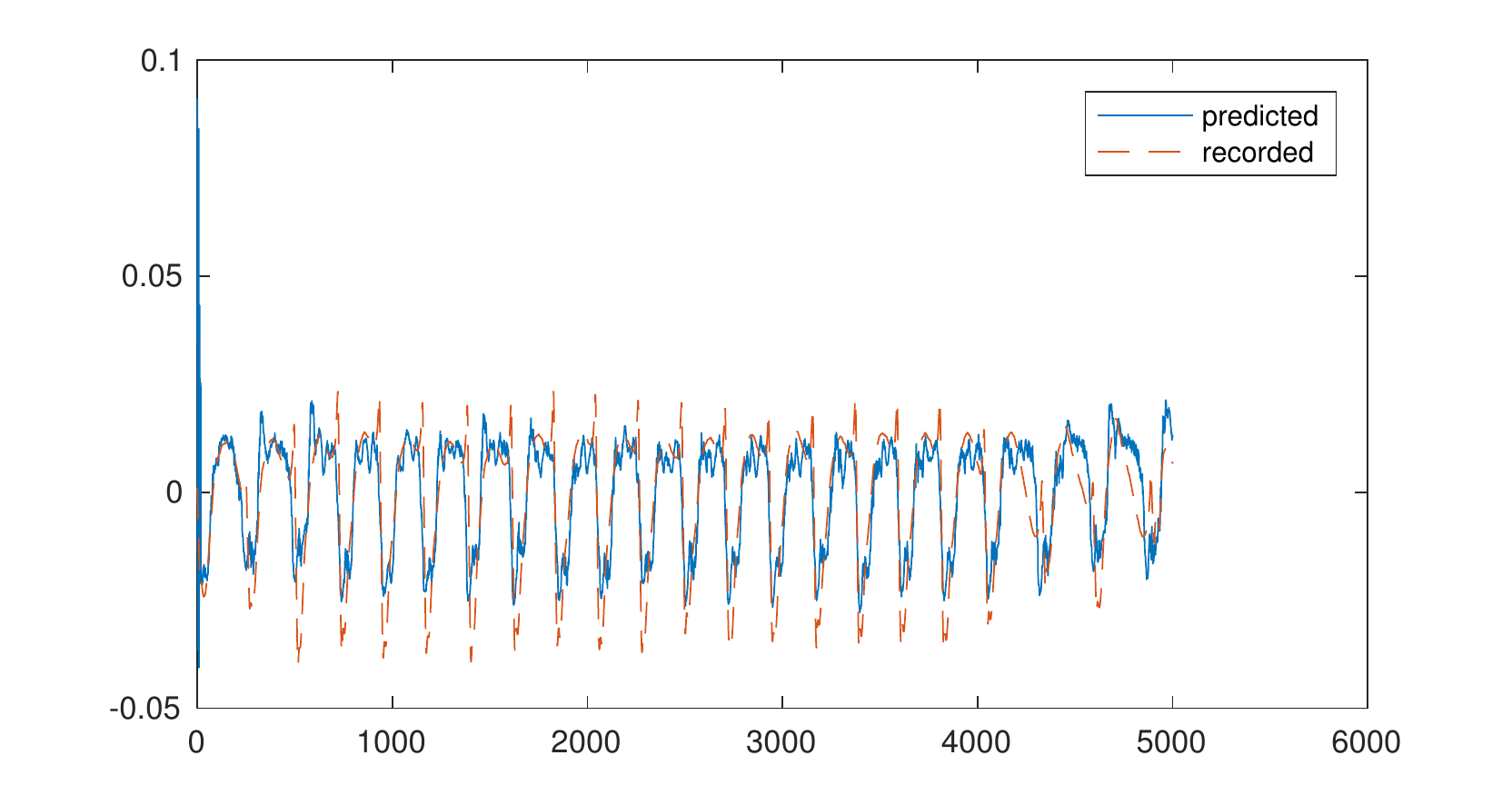}} 
  \caption{Test result Patient: 198a385, ventricular flutter,
    correlation coefficient: 0.8171, $N=5$, $d=18$.}
  \label{fig:res2c}
\end{figure}

Figures~\ref{fig:res2a}-\ref{fig:res2c} correlate the predicted output
and the patient recorded HSP output, at the right ventricular apex,
under ventricular flutter. In this case, the correlation coefficient
approach 0.9, with an average for the three patients at 0.7. These
values are very close to the ideal value of 1.0. Hence, this shows that
our TDANN approach is a good option for predicting HSP from BSP.

\begin{figure}[h!]
  \centering
  \includegraphics[scale=0.44]{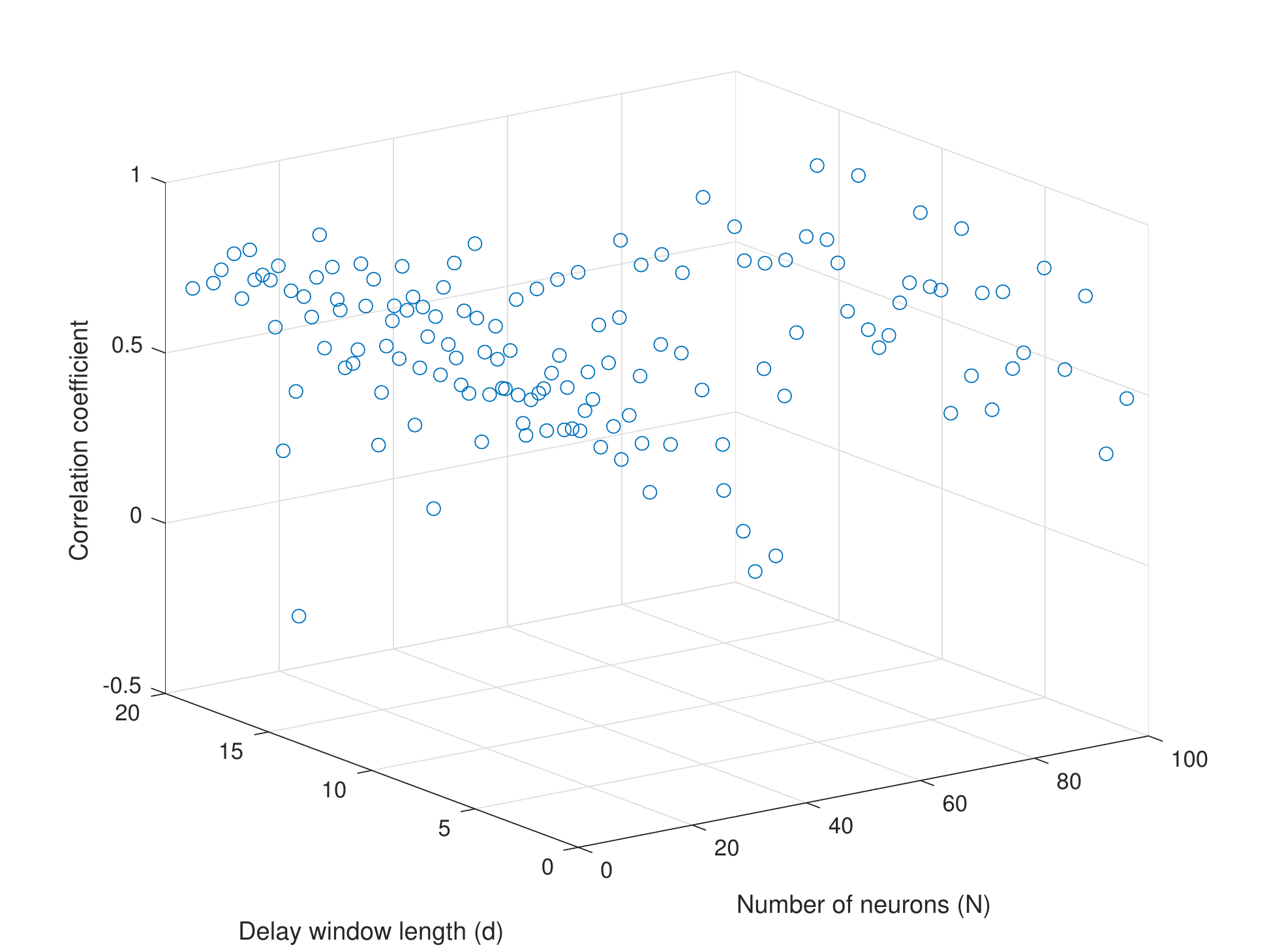}
  \caption{Affect of changing delay window size
    ($d$) and number of neurons ($N$) for the test dataset 198a385}
  \label{fig:redsd}
\end{figure}

It is worth noting that, in both cases, very few neurons and a small
delay window is needed. In fact, increasing the number of neurons and
the delay window ($d$) leads to over-fitting the output to the training
dataset (see Figure~\ref{fig:redsd}, for example). Our iterative
algorithm (Figure~\ref{fig:iteralgo}) overcomes these over-fitting
problems, by exploring the complete search space of possible solutions.

\section{Conclusions and future work}
\label{sec:conclusions}

In this paper we describe an artificial neural network approach to
predicting heart surface potentials from body surface potentials.
Predicting heart surface potentials non-invasively, can potentially be
used in the future for non-invasive cardiac diagnostics. Our primary
idea is to build time-delay neural networks for predicting single heart
surface potential from a single body surface potential. Time-delay
neural network, allows for using past values of the input body surface
potential to predict the heart surface potential. We develop an
iterative search space exploration technique to find the number of
neurons needed in the hidden layer along with the delay window size. The
prediction results are very encouraging, in that the Pearson
coefficients correlating predicted and recorded heart surface potentials
approach the ideal value under normal and diseased heart states. We have
shown the efficacy of our approach using real-world recorded patient
data.

In the future we plan to enhance the presented approach with multiple
body surface potential recordings to predict the heart surface
potentials.

\addtolength{\textheight}{-12cm}  




\end{document}